\pdfoutput=1
\documentclass[11pt]{article}

\usepackage[final]{acl}

\usepackage{times}
\usepackage{latexsym}

\usepackage[T1]{fontenc}
\usepackage[utf8]{inputenc}

\usepackage{microtype}

\usepackage{inconsolata}

\usepackage{graphicx}

\usepackage{amsmath}

\usepackage{multirow}

\title{\longtitle}

\author{
Muhammad Umair \and Vasanth Sarathy \and JP de Ruiter\\ 
Department of Computer Science\\
Tufts University\\  
Medford, Massachusetts, USA\\
\texttt{\{muhammad.umair, vasanth.sarathy, jp.deruiter\}@tufts.edu}
}

\usepackage{xspace}
\usepackage[draft]{commenting}

\newcommand{\longtitle}{Large Language Models Know What To Say But Not When To Speak\xspace}

\usepackage{commenting}
\declareauthor{vs}{Vasanth}{blue}
\declareauthor{jp}{JP}{red}

\newcommand{\hide}[1]{}

\begin{document}

\maketitle

\begin{abstract}
Turn-taking is a fundamental mechanism in human communication that ensures smooth and coherent verbal interactions. Recent advances in Large Language Models (LLMs) have motivated their use in improving the turn-taking capabilities of Spoken Dialogue Systems (SDS), such as their ability to respond at appropriate times. However, existing models often struggle to predict \emph{opportunities} for speaking — called Transition Relevance Places (TRPs) — in natural, unscripted conversations, focusing only on turn-final TRPs and not within-turn TRPs. To address these limitations, we introduce a novel dataset of participant-labeled within-turn TRPs and use it to evaluate the performance of state-of-the-art LLMs in predicting opportunities for speaking. Our experiments reveal the current limitations of LLMs in modeling unscripted spoken interactions, highlighting areas for improvement and paving the way for more naturalistic dialogue systems.

\end{abstract}

\newcommand{\turn}{U}
\newcommand{\word}[1]{w_{#1}}
\newcommand{\rword}[1]{\tilde{w}_{#1}}
\newcommand{\turnLen}{N}
\newcommand{\respLen}{M}
\newcommand{\prefix}[1]{P_{#1}}
\newcommand{\prefixes}{\mathcal{P}_\stimulus}
\newcommand{\TRP}[1]{T_{#1}}
\newcommand{\vocab}{L}
\newcommand{\TRPs}{\mathcal{T}_{\response, \stimulus}}
\newcommand{\TRPsPrediction}{\TRPs^{Predicted}}
\newcommand{\TRPsParticipants}{\TRPs^{Participants}}
\newcommand{\stimulus}{S}
\newcommand{\response}{R}
\newcommand{\wordStart}[1]{t_{\word{#1}}^s}
\newcommand{\wordEnd}[1]{t_{\word{#1}}^e}
\newcommand{\wordMid}[1]{t_{\word{#1}}^m}
\newcommand{\rwordStart}[1]{t_{\rword{#1}}^s}
\newcommand{\rwordEnd}[1]{t_{\rword{#1}}^e}
\newcommand{\rwordMid}[1]{t_{\rword{#1}}^m}
\newcommand{\interval}[2]{I_{#1#2}}
\newcommand{\intervals}[1]{\mathcal{I}_{#1}}
\newcommand{\intervalProportion}[2]{\interval{#1}{#2}^{Proportion}}
\newcommand{\participant}[1]{a_{#1}}
\newcommand{\participants}{A}
\newcommand{\Distance}{\mathcal{D}_{\stimulus}}

\section{Introduction}
\label{sec:intro}

When humans interact verbally, they avoid speaking simultaneously and take turns to speak and listen, a process essential for mutual understanding and smooth communication \citep{stivers2009universals, deRuiter2019TurnTaking}. Unlike in formal settings with pre-assigned roles, participants in everyday conversations decide when to speak or listen on a per-turn basis \citep{sacks1974simplest}. This \textit{local management system} hinges on conversationalists' ability to recognize and anticipate so-called \textit{Transition Relevance Places} (TRPs), which are points in a speaker's utterance that signal appropriate opportunities for the listener to speak. In other words, at a TRP, listeners have the opportunity, but are not obligated, to speak. Importantly, interlocutors anticipate and recognize TRPs using various lexico-syntactic, contextual, and intonational cues \citep{deRuiter2006ProjectingTheEnd, boegels2021cues}.

The ability to predict TRPs is therefore crucial for artificial conversational agents, as it enables them to take turns and provide verbal feedback signals with socially appropriate timing. Recent advances in Large Language Models (LLMs) have generated interest in improving turn-taking capabilities in Spoken Dialogue Systems (SDS) using these models \citep{Ni2021RecentAI}. Specifically, approaches like TurnGPT and RC-TurnGPT introduce probabilistic models to predict TRPs using contextual and speaker-identity information \citep{ekstedt2020turngpt, jiang-etal-2023-response}. However, most methods struggle to handle unscripted spoken interactions, often resulting in long silences or poorly timed feedback \cite{skantze2021turnreview}.

There are two critical issues with the current approaches. First is the optimistic assumption that LLMs trained predominantly on written-first language can learn the complex dynamics of spoken-first language \citep{MAHOWALD2024,umair2024predicting,liesenfeld2024rethinking}, which are distinct in structure and use (e.g. \citealt{drieman1962differences, pilan2023conversational}). A second, more fundamental issue is that, while TRPs at speaker switches can be identified unambiguously, it is challenging to clearly identify TRPs \emph{within} turns. This means that we have no `ground truth' data about these `silent' TRPs, where a listener could have responded but chose not to.

In this work, we address these issues by first developing a novel and unique empirical dataset\footnote{The dataset collected as part of this work is publicly available at: \url{https://osf.io/k5pc9/?view_only=5124d862448f4435b775d49a7b299d6d}} based on human responses that allows us to identify within-turn TRPs in natural conversation.  Second, we use this dataset to establish the baseline performance of state-of-the-art LLMs to predict within-turn TRPs. This ability is vital for future dialogue systems to appropriately time their turns, use strategically timed silences to convey social cues, and maintain conversational flow.

\section{Theoretical Background}
\label{sec:background}

When humans verbally interact with each other, they avoid speaking at the same time and instead take turns speaking and listening \cite{sacks1974simplest}. This allows them to respond sequentially to each other's utterances and facilitates mutual comprehensibility \citep{Duncan1972SomeSignals,deRuiter2019TurnTaking}. However, the alternation between speaker and listener roles in natural conversation is not predetermined (e.g., allotted time slots in court proceedings). Rather, it is \emph{locally} managed by speakers themselves on a per-turn basis \citep{stivers2009universals,Torreira2015IntonationalBoundaries}. But how can participants in conversations manage to avoid speaking at the same time, or having long silences in which they are waiting for one another? 

Conversationalists follow rules imposed by a \emph{universal} turn-taking model, proposed by \citet{sacks1974simplest} \citep[see also][]{levinson1983pragmatics, deRuiter2019TurnTaking}. This system crucially depends on the notion of the \textit{Transition Relevance Place} (TRP), which is an opportunity in the current speaker's utterance at which a listener can, but is not obligated to, take over the role of speaker. Even short feedback-like turns, such as `hmm', known as \textit{backchannels} \citep{yngve1970backchannel} or \textit{continuers} \citep{schegloff1982discourse}, are precisely timed to occur at TRPs. Importantly, TRPs are not a function of a speaker's intentions but a consequence of the turn-taking mechanism itself. This distinguishes speech at TRPs from interruptions or barge-ins, which can occur at any point in a speaker's utterance and are noticeable precisely because their timing does not meet normative expectations. 

In the turn-taking literature, a crucial distinction is made between a \textit{turn}, which is the entire contribution by one speaker, and a \textit{Turn Construction Unit} (TCU), which ends at a TRP. Since a listener is not required to speak at every TRP, a turn can consist of multiple TCUs, with potentially multiple TRPs occurring within a turn. This implies that, by definition, turn-switches can only occur at a TRP. While it is relatively straightforward to identify turn-final TRPs -- where the listener takes over -- it is challenging to reliably locate turn-medial TRPs (where the speaker continues) due to the absence of observable cues suggesting the presence of a TRP. 

To function, the turn-taking system requires listeners to not only recognize but also \textit{anticipate} the occurrence of a TRP in the current speaker's contribution \cite{riest2015anticipation}. Listeners process various turn-taking cues -- primarily lexico-syntactic \citep{deRuiter2006ProjectingTheEnd}, but also contextual and intonational cues \citep{boegels2021cues} -- incrementally to predict upcoming TRPs, ensuring that their responses are normatively timed. This anticipatory ability is essential not only for successful human interactions but also for designing artificial conversational agents capable of a) taking over the floor at the right moment and b) providing verbal feedback with correct timing.

Beyond the basic mechanisms, cultural variations also play a role in how turn-taking unfolds. While the fundamental mechanisms of turn-taking, such as the cues for taking or passing on turns \citep{stivers2009universals}, are largely universal, cultural norms can shape the timing and style of these transitions \citep{schegloff1982discourse}. Therefore, understanding both the culture-invariant and culture-specific components of turn-taking is crucial for developing dialogue systems that are not only responsive but also adaptable across diverse cultural contexts. 

\section{Related work}

Efforts to improve turn-taking in Spoken Dialogue Systems (SDS) have increasingly leveraged the linguistic capabilities of LLMs, driven by the need for these systems to manage natural unscripted interactions, particularly in multi-party settings \citep{Ni2021RecentAI}. One notable approach is TurnGPT, which introduces a probabilistic model to predict Transition Relevance Places (TRPs) using turn-shift tokens based on both contextual and speaker-identity information \citep{ekstedt2020turngpt}. An extension of this approach is RC-TurnGPT, which incorporates the predicted responses of interlocutors, conditioning predictions on upcoming linguistic content \citep{jiang-etal-2023-response}. However, these methods so far do not generalize well to corpora of \emph{unscripted} dialogue. As a consequence, current dialogue systems still tend to produce long silences and ill-timed feedback \cite{skantze2021turnreview}. 

Ablation studies on these LLMs suggest that previous linguistic content is generally sufficient for accurate prediction of turn-ends, and that the relative gain in accuracy diminishes with larger context windows—i.e., TRPs can often be predicted effectively using the local linguistic content of a turn. Additionally, although several approaches attempt to predict turn-ends using acoustic signals (e.g., \citealt{ekstedt2022much, ekstedt2022voice, inoue2024multilingual}), our work is grounded in human turn-taking literature, where linguistic cues are recognized as both necessary and sufficient for anticipating TRPs \citep{deRuiter2006ProjectingTheEnd}.

\section{Our approach}

There are two common methods for identifying TRPs in recorded conversation corpora (e.g, \citealt{godfrey1993switchboard,anderson1991hcrc,kraaij2005ami}). The first is to locate speaker changes, which are directly observable, and infer that these changes occur only at TRPs. The second is to have experts in conversation analysis manually annotate TRPs. While these methods are widely used, both have limitations. Speaker changes only account for a subset of all TRPs, as there are opportunities where a listener could respond but chooses not to i.e., within-turn TRPs \citep{threlkeld2022using}. Expert annotations, meanwhile, are subjective and do not align with the task faced by participants in real-time dialogue, who must predict TRPs instinctively `on-the-fly'. In contrast, annotators analyze conversations retroactively, without engaging in the same anticipatory processing as active participants -- leading to low ecological validity (see \citet{albert2018improving}).

To address these limitations, we designed an experiment that engaged participants in natural conversations, asking them to produce auditory responses at any location they felt it was possible to respond—not necessarily where they would have responded in everyday interactions. By having multiple participants repeat this task, we gathered a wide range of responses for the same turns. While individual responses varied, the aggregated distribution (with a sufficiently large sample size) provides a reliable indicator of within-turn TRPs, reflecting how humans identify these opportunities in real conversations.

\subsection{Collection of data on human-detected TRPs in natural turns}
\label{sec:dataset}

\begin{figure*}[t!]
    \centering
    \includegraphics[width=1\linewidth]{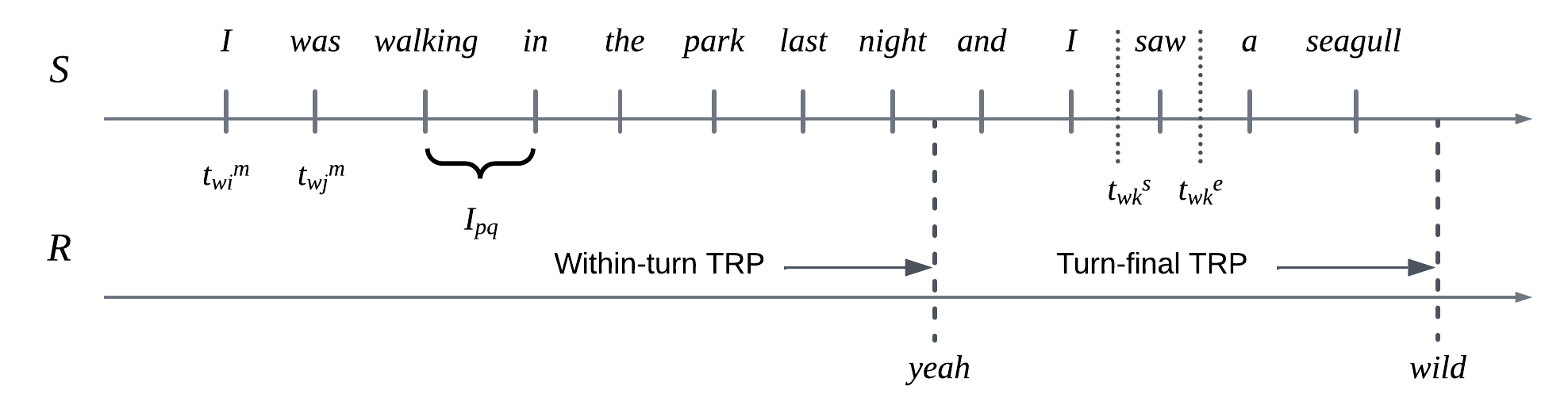}
    \caption{Participants listened to a stimulus ($S$) and produced auditory responses ($R$) to indicate their perception of TRPs. Each word in the stimulus ($\word{1},\wordStart{1},\wordEnd{1}$) and the response ($\rword{1},\rwordStart{1},\rwordEnd{1}$) has a start and end time. Intervals are between adjacent words ($\interval{p}{q}$).}
    \label{fig:formalism}
\end{figure*}

\subsubsection*{Corpus of natural conversations}

To create a reliable dataset of participants' instinctive responses to TRPs, we first collected a corpus -- named the \emph{In Conversation Corpus}\footnote{The ICC is not currently publicly available in its entirety due to restrictions by the Tufts University IRB.} (ICC) -- containing high-quality recordings of informal dialogues in American English. The ICC consists of 93 conversations, each lasting approximately 25 minutes, and each featuring pairs of undergraduate students engaged in unscripted conversations. Participants sat in sound-proofed rooms separated by a glass window and communicated using microphones and headphones, ensuring that we could record high-quality audio with complete sound isolation per speaker i.e., no cross-talk. The recordings were first automatically transcribed using GailBot \citep{umair2022gailbot}, following \emph{Jeffersonian} transcription notation, and subsequently verified by human annotators. 

From the 93 total conversations in the ICC, we initially selected 17 candidate conversations (approximately 425 minutes of talk) for further analysis. To focus specifically on participants' instinctive localization of within-turn TRPs, we further filtered these conversations to select 55 turns that we suspected contained at least two TCUs. Ultimately, this resulted in 28.33 minutes of talk being used in the data collection reported below. 

We selected the ICC over publicly available dialogue corpora to ensure more natural and diverse turn-taking behaviors in our dataset. Although open-source datasets are well-annotated and widely studied, their data collection methods often limit the range and authenticity of conversational behaviors exhibited \citep{reece2023Candor}.

\subsubsection*{Empirical collection of estimated TRP locations.}
\label{sec:trpCollection}

To empirically determine the locations of TRPs, we created two mutually exclusive lists of \emph{stimulus turns} from the filtered subset of the ICC (55 turns; 28.33 minutes of talk), ensuring that each turn was assigned to only one list. To mitigate potential ordering effects, we generated two additional lists in which the stimulus turns (see Figure \ref{fig:formalism}) were presented in reverse order (i.e., the last turn appeared first, and so on). In total, we had four \emph{stimulus lists}, each approximately 15 minutes long.

We recruited 118 native English speakers as participants\footnote{This study was approved by the Tufts University IRB (ID = STUDY00003236). Participants were undergraduates and were compensated as per IRB regulations.}, none of whom were experts in the turn-taking literature. Each participant was randomly assigned to one of the four stimulus lists and asked to verbalize brief backchannels (e.g., `hmm', `yes') whenever they felt it was appropriate. Each participant's responses were recorded on individual audio channels, synchronized with the stimulus audio to maintain clarity and separation.

\begin{figure*}[hbt!]
    \centering
    \includegraphics[scale=0.6]{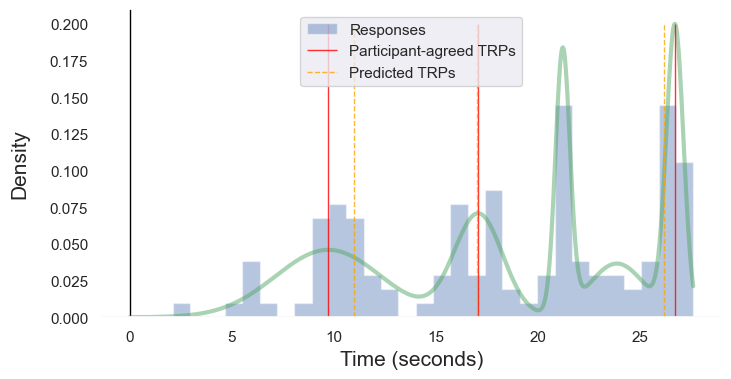}
    \caption{Distribution of participant responses, the times at which participants agreed a TRP occurred, and model predictions of TRPs for a single stimulus $\stimulus$. The dotted lines indicate that each participant-agreed TRP has some associated variance. The responses are binned between the temporal midpoint of words (see Section \ref{sec:task}).} 
    \label{fig:exampleStimulus}
\end{figure*}

We used the phonetic analysis software Praat \citep{boersma2001speak} and ELAN \citep{wittenburg-etal-2006-elan} to \emph{manually} locate the onset of each backchannel response across all participants.  This allowed us to ensure that we used precise timing for words and did not accidentally consider other types of speech (e.g., in-breaths, out-breaths, laughter etc.) as responses. Since two of the lists were reversals of the originals, we merged the participant responses from these reversed lists with those from the original lists for analysis. On average, 59 participants responded to each stimulus turn, multiple times if they perceived multiple TRPs, resulting in an average of 159 responses per stimulus turn (see Table \ref{tab:ListsSummary}). This allows us to estimate both the likelihood of perceiving a TRP at a specific location and the distribution of those estimated response locations (see Figure \ref{fig:exampleStimulus}). Refer to Appendix \ref{app:procedures} for further details on the processing of stimulus lists.

\begin{table}[htp!]
    \centering
    \renewcommand{\arraystretch}{1.2}
    \setlength{\tabcolsep}{10pt} 
    \resizebox{\columnwidth}{!}{
        \begin{tabular}{p{4.0cm}ll}
            \hline
             & \multicolumn{2}{c}{\textbf{Stimulus Lists}}\\
            \textbf{Metric} & \textbf{List 1} & \textbf{List 2} \\ 
            \hline 
            List duration (s) & 846.3 & 853.5\\ 
            \# of words  & 2558 & 2693\\ 
            \# of participants & 60 & 58\\ 
            \# of stimuli  & 28 & 27\\ 
            Avg. stimulus duration (s) & 30.5 & 31.7\\ 
            \# words per stimulus  & 91.3 & 99.7 \\ 
            Avg. \# of responses per stimulus & 156 & 162\\    
            \hline
        \end{tabular}
    }
    \caption{Participants listened to two stimulus lists and their reversals, each containing multiple turns. They indicated within-turn TRPs using brief auditory backchannels. Responses from the original and reversed lists were merged for analysis. The table summarizes statistics for each list. Note that \# refers to number with the duration in seconds.}
    \label{tab:ListsSummary}
\end{table}

\subsection{Within-Turn TRP Prediction Task}
\label{sec:task}

\hide{
This section details your approach to the problem. 
\begin{itemize}
    \item Please be specific when describing your main approaches. You may want to include key equations and figures (though it is fine if you want to defer creating time-consuming figures until the final report).
    \item Describe your baselines. Depending on space constraints and how standard your baseline is, you might do this in detail or simply refer to other papers for details. Default project teams can do the latter when describing the provided baseline model.
    \item If any part of your approach is original, make it clear. For models and techniques that are not yours, provide references.
    \item If you are using any code that you did not write yourself, make it clear and provide a reference or link. 
    When describing something you coded yourself, make it clear.
\end{itemize} 
}

\subsection*{Preprocessing Multi-channel Audio Data}

Since we are evaluating the ability of LLMs to recognize TRPs, we require a principled method to formalize our experimentally collected dataset by converting the audio data from two synchronized channels (stimulus and response) into a structured format suitable for analysis. From the recorded audio, we extract words along with their precise timing information, including start and end times. To ensure high accuracy, these timing details were manually annotated to the nearest tenth of a second for both the stimulus and participant responses.

Formally, we define a single stimulus $\stimulus = \langle (\word{1},\wordStart{1},\wordEnd{1}), \ldots, (\word{\turnLen},\wordStart{\turnLen},\wordEnd{\turnLen}) \rangle$ of length $\turnLen$ as a sequence of words $\word{i}$, where each word belongs to a fixed vocabulary $\vocab$, such that $\forall \word{i} \in \stimulus, \word{i} \in \vocab$. The stimulus $\stimulus$ also includes start ($\wordStart{i}$) and end ($\wordEnd{i}$) times for each word.  Participant responses are similarly defined as $\response = \langle (\rword{1},\rwordStart{1},\rwordEnd{1}), \ldots, (\rword{\respLen},\rwordStart{\respLen},\rwordEnd{\respLen}) \rangle$. Further, we calculate the temporal midpoint of each word as $\wordMid{i} = (\wordStart{i} + \wordEnd{i})/2$, and use these midpoints to create intervals, $\interval{i}{j}, 1 \leq i,j \leq \turnLen, j = i+1$, between words. Using the temporal midpoint provides a more reliable estimate for determining whether a response is most reasonably associated with the preceding word.

We also define a binary random variable $\TRP{i} \in \{0,1\}$ for each interval $\interval{i}{j}$ indicating the occurrence (1) or absence (0) of a TRP after word $\word{i}$. The vector $\TRPs = \langle \TRP{1}, \ldots, \TRP{\turnLen} \rangle$ subsequently acts as a collection of binary indicators representing whether a TRP occurred in each interval $\interval{i}{j}$ of a stimulus $\stimulus$.

Finally, $\TRPsParticipants$ represents a binary indicator of whether participants agreed that a TRP had occurred in each interval of a stimulus $\stimulus$. We determine participant agreement by calculating the proportion of participant responses $\intervalProportion{i}{j}$, based on their start times ($\rwordStart{i}$), that fall within each interval $\interval{i}{j}$. We consider a TRP to have occurred if the proportion of responses for an interval exceeds a predefined threshold $\tau \in [0,1]$, i.e., $\intervalProportion{i}{j} > \tau$ (we used $\tau = 0.3$). Note that the choice of $\tau$ is crucial, as it directly impacts $\TRPsParticipants$: a larger $\tau$ requires a higher level of participant agreement for an interval to be marked as containing a TRP, while a smaller $\tau$ allows for a more relaxed consensus.

\subsection*{Task Definition}

Broadly, the inference task can be defined as identifying between 0 and $\turnLen$ TRPs in a stimulus $\stimulus$. However, it is important to consider that humans do not process entire turns as complete units; rather, we incrementally process speech and decide on the existence of TRPs at each point in time. To replicate this incremental processing in the inference task, we define a prefix $\prefix{i} = \langle \word{1},\ldots, \word{i}\ \rangle$ as a sequence of words from the first to the $i^{th}$ word, such that $\forall \word{i} \in \prefix{i}, \word{i} \in \stimulus$. We further define $\prefixes$ as the set of all prefixes for a stimulus turn $\stimulus$, with $|\prefixes| = \turnLen$

\newtheorem{definition}{Definition}[section]

\begin{definition}[Within-turn TRP Prediction]
\label{def:within_turn_trp_task}
Given a stimulus $\stimulus$, and the set of all prefixes $\prefixes$, determine $\TRPsPrediction$, where each $\TRP{i} \in \TRPsPrediction$ occurs after each of the prefixes $\prefix{i} \in \prefixes$. 
\end{definition}

Definition \ref{def:within_turn_trp_task} allows us to decompose each stimulus turn $\stimulus$ into a set of binary string classification tasks. Notably, we assume that the value of $\TRP{i}$ is independent of all prior TRP determinations, $\TRP{1}, \ldots, \TRP{i-1}$. While TRP determinations depend on multiple factors, in this paper, we focus solely on conditioning TRP determinations on the linguistic information provided by preceding words. See \citet{deRuiter2006ProjectingTheEnd} and \citet{riest2015anticipation} for experimental evidence that linguistic content is sufficient for TRP prediction.

\section{Evaluation Metrics}
\label{sec:measures}

\subsection*{Classification Metrics}

We can evaluate the performance of a model for the within-turn TRP prediction task (see Definition \ref{def:within_turn_trp_task}) by comparing its predictions $\TRPsPrediction$ against the participants' indications of TRPs $\TRPsParticipants$.  It is important to note the imbalance inherent in  the data i.e., intervals that contain TRPs are much less frequent those that do not. In this case, we cannot use accuracy since a model that simply predicts the majority class for all intervals will have achieve a high value. Instead, the F1 score i.e., the harmonic mean of precision and recall, is well suited since it emphasizes models that perform well in identifying intervals that contain TRPs ($\TRP{i} = 1$), which are the vast minority of intervals.

\subsection*{Free-Marginal Multirater Kappa}

Multirater Kappa statistics are often used in medical and behavioral sciences as a measure of agreement over chance between multiple raters \cite{artstein2008inter}. There are a number of benefits to using Kappa in the context of our work. First, most LLMs, especially smaller ones, lack consistency over multiple predictions generated with the same prompt. Additionally, since most state-of-the-art LLMs do not provide direct access to probability distributions, the kappa statistic can be used to directly compare multiple responses from the same model. In fact, it can also be used to assess agreement between groups of models \cite{tang2024tofueval}. Second, kappa is a measure of \emph{reliability}, but not validity. It might be the case that groups of LLMs may agree with each other, but not with human participants. Therefore, the kappa statistic offers a way to compare predictions of LLMs to human evaluators \cite{wang2024prompt}. This is especially important when considering TRPs since the subjectivity of turn-taking decisions may lead to disagreement between raters (LLMs or humans), but might not necessarily indicate an incorrect prediction. 

Fleiss' Kappa is typically used when there are multiple raters assessing a nominal variable \cite{fleiss1971measuring}. It assumes that the $n$ raters know a priori the number of cases $N$ that must be assigned to each category $K$. However, this assumption is not valid in our task, which consists of raters (the participants and the models) attempting to assign binary TRP categories across a number of cases (each interval is a case). Here, the rater does not know a priori the number of TRPs that occur in a specific stimulus. When this assumption does not hold, the value of Fleiss' kappa can change significantly based on the distribution of cases in each category, even when all other variables are held constant. \citet{randolph2005free} proposed a kappa measure (see Equation \ref{eq:randKappa}) that resolves this issue and does not make any assumptions about the number of cases in each category (number of TRPs in our case). 

\begin{align}
\label{eq:randKappa}
    k_{free} = \frac{[\frac{1}{Nn(n-1)} \sum_{i=1}^{N}\sum_{j=1}^{K} n_{ij}^2 - Nn] - \frac{1}{k}}{1 - \frac{1}{k}}
\end{align}

\begin{figure}[htp!]
    \centering
    \includegraphics[scale=0.6]{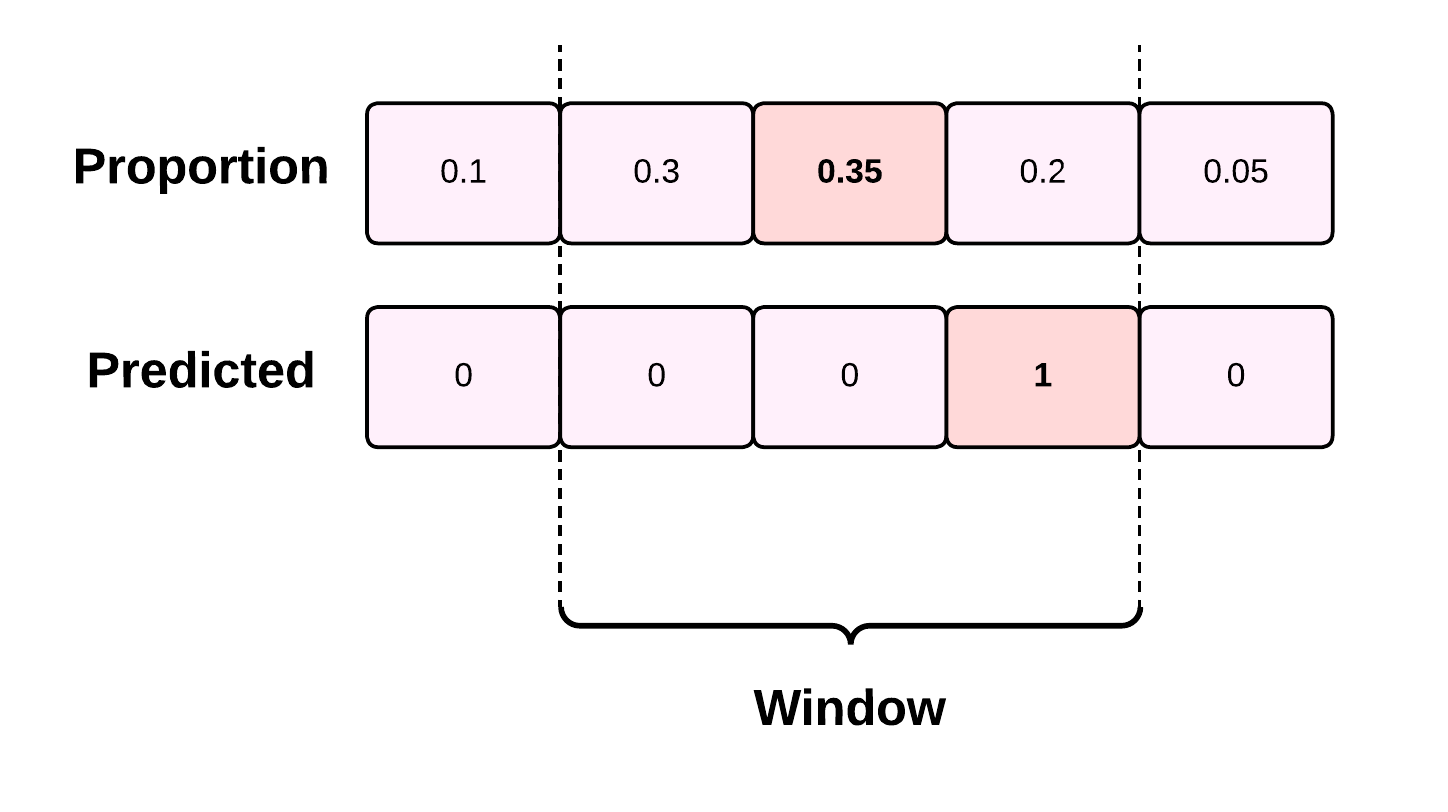}
    \caption{Example of participant response proportions and corresponding model predictions in each interval of a sample stimulus $\stimulus$. In this example, $\tau = 0.3$, which means that there is one one interval in which participants agree that a TRP has occurred. Due to variance in human indications of TRP locations, we may consider a correct prediction to have occurred within some window of the participant-agreed TRP.} 
    \label{fig:exampleVectors}
\end{figure}

We calculate two variants of the Kappa statistic. The first, $k_{free}^{all}$, computes the kappa statistic across all intervals, as previously described. However, since our primary focus is on intervals where participants agreed that a TRP occurred—representing only a small portion of the total intervals—considering all intervals may result in an inflated kappa value, falsely indicating a high level of agreement. To address this, we also calculate $k_{free}^{true}$, which specifically evaluates the kappa statistic for intervals where a TRP was present. Moreover, $k_{free}^{true}$ accounts for the density of participant responses by marking a model prediction as "correct" if it falls within a defined window around an interval where participants agreed there was a TRP (see Figure \ref{fig:exampleVectors}).

\subsection*{Temporal Distance Metrics}

Let $d_{i,j}^{\stimulus} \in {1, \ldots, N}$ (see Equation \ref{eq:minDist}) represent the minimum absolute distance, in terms of the number of intervals, between an interval where a response was predicted ($\TRP{i}^{Predicted} = 1$) and the closest interval in which participants agreed that a TRP occurred ($\TRP{j}^{Participants} = 1$). Furthermore, let $\Distance = \langle d_{i,j}^{\stimulus}, \ldots, d_{p,q}^{\stimulus} \rangle$ be a vector of these distances, with $|\Distance| = K$, where $K$ is the total number of predicted TRPs.

As previously discussed (see Section \ref{sec:task}), we consider an interval $\interval{i}{j}$ to contain a TRP ($\TRP{j}^{Participants} = 1$) if the proportion of participants that responded in that interval exceeds a predefined threshold ($\intervalProportion{i}{j} > \tau$). However, a model's prediction may not align perfectly with the exact interval $\interval{i}{j}$ where participants agreed that a TRP occurred. Instead, the prediction may fall in a nearby interval that still lies within an acceptable range, given the inherent variance in participant responses regarding the precise location of TRPs (e.g., \citet{templeton2022fast}). 

\begin{align}
    d_{i,j}^{\stimulus} &= min(|i-j|) \forall j \in \TRP{j}^{Participants} = 1
    \label{eq:minDist}
\end{align}

Therefore, we define two measures of temporal distance between the predicted TRP and the closest participant-agreed TRP location: the Normalized Mean Absolute Error (NMAE) and the Normalized Mean Square Error (NMSE). The NMAE provides a linear measure of distance, whereas the NMSE offers a quadratic measure.

\begin{align*}
    \textit{NMAE} &= \sum_{i=1}^{|\Distance|} d_{i,j}^{\stimulus}\\
    \textit{NMSE} &= \sum_{i=1}^{|\Distance|} (d_{i,j}^{\stimulus})^2
\end{align*}

However, these simple measures do not account for the \emph{density} of responses around an interval where a TRP occurred. For instance, if the density of responses near this interval is high, it may be reasonable to expect that a TRP could also occur in the neighboring intervals. To incorporate this, we employ a windowed approach to calculate response density. For each interval where a participants agreed a TRP occurred ($\intervalProportion{i}{j} > \tau$), we center a window of size $W$ on that interval. The density of responses is then defined as the proportion of participant responses within the entire window, which we use to compute the \emph{density-adjusted} measure $\textit{NMAE}_{DA}$.

\begin{align*}
    \textit{Density}_S(\interval{i}{j}, W) &= \sum_{l=\frac{-W}{2}}^{\frac{W}{2}} \intervalProportion{i+l}{j+l}\\ 
    \textit{NMAE}_{DA} &= \sum_{i=1}^{|\Distance|} \frac{d_{i,j}^{\stimulus}}{\textit{Density}_S(\interval{i}{j},W)}
\end{align*}

\section{Experiments and Results}

We employed several state-of-the-art LLMs to perform the within-turn TRP prediction task described in Section \ref{sec:task}. Our focus was on models pre-trained on diverse datasets, some of which were explicitly designed with capabilities for spoken interaction.

To adapt LLMs for downstream tasks, two main strategies are commonly used: fine-tuning and in-context learning (ICL). Fine-tuning involves updating the weights of a pre-trained model to specialize it for a particular task, resulting in a single model tailored for that task. This approach is advantageous because it can accommodate training sets of any size, often leading to significant performance improvements. However, most state-of-the-art LLMs are not available for direct fine-tuning due to restricted open-source access and instead are accessible only through public APIs \cite{liesenfeld2024rethinking}.

Given these limitations, we employed in-context learning (ICL) as our adaptation strategy. Unlike fine-tuning, ICL does not require modifying model weights. Instead, it adapts the model to a specific task by using task demonstrations provided through prompts. However, it is crucial to note that ICL is highly sensitive to the formulation of these prompts, and optimizing prompts requires careful consideration and specific strategies \cite{chang2024efficient}. 

We tested each model under two prompting conditions: expert and participant. In the expert condition, the model was provided with theoretical background on TRPs, similar to what an expert annotator might know. In the participant condition, the model was given a version of the instructions that the human participants received. These two prompting conditions explore how the level of provided information affects the model's performance on the TRP prediction task.

\begin{table*}[htp!]
    \centering
    \renewcommand{\arraystretch}{1.2}
    \small
    \begin{tabular}{ll|cccccccc}
        \hline
        \textbf{Model} & \textbf{Condition} & \textbf{Precision} & \textbf{Recall} & \textbf{F1 Score} & $\mathbf{k_{free}^{all}}$ & $\mathbf{k_{free}^{true}}$ & \textbf{\textit{NMAE}} & \textbf{\textit{NMSE}} & $\mathbf{\textit{NMAE}_{DA}}$ \\
        \hline
        \multirow{2}{*}{GPT-4 Omni} & Participant & \textbf{0.153} & 0.153 & \textbf{0.152} & \textbf{0.891} & \textbf{0.325} & 0.286 & \textbf{3.140} & \textbf{11.280} \\
                               & Expert      & 0.122 & 0.185 & 0.147 & 0.860 & 0.201 & 0.253 & 5.360 & 16.560 \\
        \hline
   
        \multirow{2}{*}{Phi3:3.8b} & Participant & 0.034 & 0.923 & 0.067 & -0.671 & -0.417 & 0.192 & 5.189 & 16.430 \\
                               & Expert      & 0.031 & 0.083 & 0.045 & 0.779 & 0.001 & 0.251 & 8.648 & 21.640 \\
        \hline
        \multirow{2}{*}{Phi3:14b} & Participant & 0.035 & 0.326 & 0.063 & 0.374 & -0.157 & 0.202 & 6.28 & 18.060 \\
                               & Expert      & 0.039 & 0.057 & 0.046 & 0.845 & 0.137 & 0.232 & 5.091 & 16.920 \\
        \hline
        \multirow{2}{*}{Gemma2:9b} & Participant & 0.028 & 0.285 & 0.052 & 0.322 & -0.088 & 0.224 & 8.059 & 20.770 \\
                               & Expert      & 0.022 & 0.178 & 0.039 & 0.441 & -0.087 & 0.239 & 8.784 & 22.180 \\
        \hline
        \multirow{2}{*}{Gemma2:27b} & Participant & 0.033 & 0.490 & 0.063 & 0.034 & -0.387 & 0.194 & 5.26 & 16.650 \\
                               & Expert      & 0.039 & 0.307 & 0.068 & 0.459 & -0.232 & 0.206 & 5.79 & 17.560 \\
        \hline
        \multirow{2}{*}{Llama3.1:8b} & Participant & 0.014 & 0.082 & 0.025 & 0.618 & -0.106 & 0.265 & 9.815 & 24.320 \\
                               & Expert      & 0.020 & 0.077 & 0.032 & 0.692 & -0.071 & 0.268 & 9.947 & 24.420 \\
        \hline
        \multirow{2}{*}{Mistral:7b} & Participant & 0.033 & \textbf{0.804} & 0.064 & -0.517 & -0.413 & 0.194 & 5.168 & 16.510 \\
                               & Expert      & 0.037 & 0.266 & 0.065 & 0.498 & -0.222 & \textbf{0.190} & 5.136 & 16.110 \\
        \hline
    \end{tabular}
    \caption{Measures of performance for multiple models on the within-turn TRP prediction task (see Section \ref{sec:task}) in both participant and expert contexts. The results indicate that, despite being the strongest performer overall, GPT-4 Omni still performs poorly on the task.}
    \label{tab:results}
\end{table*}

Table \ref{tab:results} shows the performance of multiple language models, including GPT-4 Omni, Phi3, Gemma2, Llama3.1, and Mistral, on the within-turn TRP prediction task averaged across all stimulus lists (see Section \ref{sec:dataset}). We focus on GPT-4 Omni because it is the best overall performer, setting the benchmark for this challenging task, despite its significant shortcomings. While other models, like Mistral:7b and Phi3:14b, show strengths in specific metrics—such as lower NMAE (0.190) and favorable NMSE (5.091) in expert conditions—these are limited to isolated scenarios, and overall performance across metrics like precision, recall, and F1 score remains inferior to GPT-4 Omni. 

Overall, the performance of the best performing model reveals significant shortcomings. First, the model exhibits low precision (0.137) and recall (0.169), leading to a low F1 score (0.151), indicating frequent false positives and missed TRPs. While the kappa statistic across all intervals ($k_{free}^{all}$ = 0.876) suggests good general agreement, the much lower kappa for participant-agreed TRP intervals ($k_{free}^{true}$ = 0.263) highlights difficulties in accurately identifying participant-agreed TRPs. The NMAE (0.263) and NMSE (4.248) metrics further indicate substantial deviations between intervals where the model predicted TRPs to the closest participant-agreed TRP. The high density-adjusted NMAE ($\textit{NMAE}_{DA}$ = 13.92) highlights even greater errors when considering the density of participant responses near intervals in which TRPs occurred. 

There are also differences between the participant and expert conditions. The expert condition yielded higher precision (0.147) compared to the participant condition (0.126), indicating more accurate identification of TRPs. The expert condition also achieved higher recall (0.185 vs. 0.153), suggesting a better ability to detect intervals in which TRPs occur. The F1 score, balancing precision and recall, was slightly higher in the expert condition (0.164) than in the participant condition (0.138). Kappa statistics also showed variability: $k_{free}^{all}$ was higher for participants (0.891 vs. 0.860), reflecting stronger overall agreement, while $k_{free}^{true}$ was higher for participants (0.325 vs. 0.201), indicating better performance in correctly identifying participant-agreed TRPs. Error metrics further demonstrated that the expert condition had lower NMAE (0.253 vs. 0.286) but higher NMSE (5.36 vs. 3.135) and significantly greater density-adjusted NMAE ($\textit{NMAE}_{DA}$ = 16.56 vs. 11.28). These results suggest that while the expert prompts provided more theoretical accuracy, the participant prompts offered more practical relevance and alignment with true TRPs. 

\section{Discussion}
\label{sec:discussion}

Half a century of research on turn-taking has demonstrated that humans rely on various cues to achieve rapid and seamless turn-transitions in natural conversation by accurately predicting upcoming TRPs. This ability is crucial for minimizing response delays and avoiding overlapping speech, both of which are interactionally significant \citep{sacks1974simplest,deRuiter2006ProjectingTheEnd,levinson2015timing}. Poorly timed turn-taking can negatively affect how utterances are interpreted; for example, longer response delays often signal reluctance or hesitation to deliver a dispreferred response \cite{deRuiter2019TurnTaking,kendrick2015timing}. Current spoken dialogue systems (SDS), however, struggle to replicate human-like turn timing, resulting in reduced user satisfaction and diminished communicative effectiveness \cite{skantze2021turnreview}. 

State-of-the-art LLMs, pre-trained on large and diverse datasets, are well-suited for leveraging linguistic information—which has been shown to be sufficient for predicting opportunities for speech in humans—and increasingly, multimodal information, to perform a range of spoken language tasks \cite{ekstedt2020turngpt,jiang-etal-2023-response,jiang2023makes}. However, contrary to expectations, we find that the LLMs we tested underperform across multiple measures on a simple binary prediction task to identify within-turn TRPs when using In-Context Learning (ICL) as the adaptation strategy. This holds true even when providing essential background context through various prompts (expert versus participant). These findings point to a major issue: LLMs are currently unable to effectively utilize their extensive linguistic knowledge for unscripted turn-taking in spoken interaction. This limits their application in dialogue systems by preventing these systems from accurately anticipating opportunities for speaker transitions.

Our work attempts to advance the performance of LLMs for turn-taking in spoken interaction. First, by empirically demonstrating that current LLMs struggle with TRP prediction despite their extensive pre-training, we expose a critical bottleneck that needs to be addressed. Second, we provide evidence that high performance on written-language benchmarks does not necessarily translate to high performance on spoken language tasks, emphasizing the need for specialized evaluation in conversational settings. Third, we contribute a specialized dataset containing empirical, on-the-fly human judgments on where TRPs occur in natural conversation. This dataset is a valuable resource for the NLP research community, offering opportunities for targeted fine-tuning and evaluation of LLMs, and enabling the development of models that more closely replicate human conversational behavior.

\section{Conclusion}

Even though Large Language Models show impressive performance on a range of challenging language-related tasks, it is as yet unclear whether they can be employed for determining when they can start producing their turn in spoken dialogue at a socially appropriate time. This would require them to have human-level ability to predict Transition Relevance Places, locations in speaker's contribution where they may take over the turn and start speaking. To test this ability in state-of-the-art LLMs, we collected data from humans that perform this task on-the-fly, and compared the performance of the LLMs with that of the human participants. It turned out that the performance of selected LLMs on this task was far below the level of that of the human participants. Apparently, the pre-training of LLMs on vast amounts of written data was not sufficient to generalize to this particular task. Possible causes for the disappointing performance could be that we haven't found the optimal prompts, and/or that the models would either need more spoken dialogue input during pre-training, or explicit fine-tuning on spoken dialogue data. Either way, the dataset that we have developed will allow researchers in the area of human-machine turn-taking to explore ways to improve the models' performance on this crucial task.

\section{Limitations}
\label{sec:future}

We acknowledge several limitations in our work. First, the models we used had access only to linguistic information, i.e., the words of a stimulus, whereas human participants had access to both prosodic and linguistic cues. Although humans can predict TRPs using only lexico-syntactic information \cite{deRuiter2006ProjectingTheEnd}, computational models often perform better with multi-modal inputs \cite{roddy2021neural, kurata2023multimodal}. Despite this, our focus on text-only models is grounded in turn-taking literature, which establishes linguistic cues as necessary and sufficient for humans to anticipate TRPs \cite{deRuiter2006ProjectingTheEnd}. Evaluating LLMs with only linguistic input was an essential step to determine whether they could replicate this human ability, particularly in spoken language contexts. Existing text-based models, such as TurnGPT and RC-TurnGPT, have struggled to generalize across diverse conversations, further emphasizing the need to isolate linguistic factors in our study. Future work should explore whether adding acoustic information can improve LLM performance on the TRP prediction task.

Second, we evaluated LLM predictions solely against participant responses from the ICC, a dataset specifically designed to capture naturalistic conversational behaviors. While we chose the ICC to avoid the limitations inherent in other corpora, it is essential to replicate our findings with commonly used dialogue datasets (e.g., Switchboard, ICSI, AMI, and SpokenWoz) to verify the broader applicability of our approach. This replication, however, is resource-intensive, as most of these corpora predominantly contain annotations for between-turn TRPs and lack detailed within-turn TRP annotations.

Third, we used In-Context Learning (ICL) as a task adaptation strategy for TRP prediction because fine-tuning was not feasible, given the restrictions on modifying the LLMs used in this work \cite{liesenfeld2024rethinking}. Although ICL has shown promise on certain tasks \cite{chang2024efficient}, its performance is highly sensitive to prompt design \cite{wei2023larger}. It is possible that we may not have fully optimized our prompts, and it remains unclear how best to engineer prompts for the within-turn TRP prediction task. Future research should explore not only the potential benefits of fine-tuning but also improved prompt engineering strategies to enhance model performance.

Finally, although LLMs can match human performance in qualitative coding tasks and provide justifications for their decisions \cite{dunivin2024scalable}, their reasoning often diverges from human reasoning \cite{bao2024llms}. While our incremental binary labeling task allows us to track the LLMs' reasoning for TRP occurrences, we did not analyze the reported reasoning in this study. Future research should focus on analyzing these reasoning patterns, as they could offer valuable insights for designing more effective prompts to improve LLM performance.

\section{Ethical Impact Statement}

Value alignment is a key concern shared by researchers and end-users of large language models. Being able to understand and model the values and normative expectations of not only the contents of speech, but the underlying communicative process itself is important to reduce the risk of misunderstandings, false attributions, and unmet normative expectations. Our work attempts to mitigate these shortcomings and provide the basis for understanding these normative nuances in communicative behavior. Our goal in releasing our corpus and these findings is to facilitate and further research in this domain. We hope to continue exploring the challenges in modeling turn-taking and evaluating the performance of large language models so as to highlight the strengths and weaknesses of using LLMs for spoken dialogue systems to researchers and practitioners.

\section{Acknowledgements}

This research was supported in part by Other Transaction award HR00112490378 from the U.S. Defense Advanced Research Projects Agency (DARPA) Friction for Accountability in Conversational Transactions (FACT) program. 

We would also like to acknowledge Grace Hustace for her contributions to data collection and processing, and Dr. Julia Mertens for her input during early-stage discussions, both of whom are affiliated with the Human Interaction Lab at Tufts University. 

\bibliography{refs}

\appendix

\section{Stimulus Preparation Procedure}
\label{app:procedures}

To prepare the stimulus lists used in this study, we first randomly sampled approximately 60\% of the conversations from the ICC, which comprises 93 conversations in total. We selected 17 conversations ( 425 minutes of talk) that met specific criteria for eliciting participant responses. These criteria included minimal background noise, no recording artifacts, and no cross-talk, ensuring that these factors would not influence participant responses.

From each selected conversation, we isolated a single audio channel representing the speech of one speaker and used ELAN to manually mark the start and end of selected turns. Turns were chosen if they were judged, by an expert annotator, to contain at least two TCUs, thus ensuring the presence of at least one within-turn TRP. The segmentation process involved two stages: an initial pass to mark provisional turn boundaries, followed by a second pass to verify and refine these boundaries. Ambiguous segments, such as those with extended pauses or unclear speaker intent, were excluded to ensure the quality of the stimuli. Ultimately, we selected 55 turns, totaling 28.33 minutes of speech.

Finally, the segmented turns were organized into stimulus lists, with each turn separated by an audible beep to indicate the start of a new turn. To minimize potential ordering effects, the turns within each list were arranged in random order. In total, four stimulus lists were generated: two with the original turn sequences and two with the sequences in reverse order. Each of the four lists was approximately 15 minutes long.

\end{document}